\documentclass[letterpaper, 10 pt, conference, preprint]{ieeeconf}  

\IEEEoverridecommandlockouts                              

\usepackage[normalem]{ulem}                                 
\usepackage{url}
\usepackage{amsmath}
\usepackage{graphicx}
\usepackage{subcaption}
\usepackage{xcolor}
\usepackage{comment}

\title{ \bf
\textbf{AuNa}: Modularly Integrated Simulation Framework for\\ Cooperative \underline{Au}tonomous \underline{Na}vigation 
}

\author{Harun Teper$^1$, Anggera Bayuwindra$^3$, Raphael Riebl$^4$, Ricardo Severino$^5$, Jian-Jia Chen$^1$, Kuan-Hsun Chen$^2$
\\
$^1$TU Dortmund University, Germany, $^2$University of Twente, the Netherlands, $^3$TU Eindhoven, the Netherlands\\
$^4$Technische Hochschule Ingolstadt, Germany, $^5$Polytechnic Institute of Porto, Portugal
} 

\begin{document}

\maketitle
\thispagestyle{plain}
\pagestyle{plain}

\begin{abstract}
In the near future, the development of autonomous driving will get more complex as the vehicles will not only rely on their own sensors but also communicate with other vehicles and the infrastructure to cooperate and improve the driving experience.
Towards this, several research areas, such as robotics, communication, and control, are required to collaborate in order to implement future-ready methods.
However, each area focuses on the development of its own components first, while the effects the components may have on the whole system are only considered at a later stage.
In this work, we integrate the simulation tools of robotics, communication and control namely ROS2, OMNeT++, and MATLAB to evaluate cooperative driving scenarios.
The framework can be utilized to develop the individual components using the designated tools, while the final evaluation can be conducted in a complete scenario, enabling the simulation of advanced multi-robot applications for cooperative driving.
Furthermore, it can be used to integrate additional tools, as the integration is done in a modular way.
We showcase the framework by demonstrating a platooning scenario under cooperative adaptive cruise control (CACC) and the ETSI ITS-G5 communication architecture.
Additionally, we compare the differences of the controller performance between the theoretical analysis and practical case study.
\end{abstract}

\section{Introduction}

In recent years, several new features for vehicles have been deployed to aid drivers, e.g., automatic parking, lane keeping assistance and autonomous driving systems for highways and low traffic areas.
While these methods only rely on the capabilities of the vehicle itself, future vehicles will also use cooperative methods to enhance the driving experience.
Although the standards for cooperative driving have been defined since 2013~\cite{c-its}, modern vehicles do not yet include these functionalities, as they introduce interactions between the communication, navigation and control systems of the vehicle.
Such features have to be assessed and evaluated before being integrated into the vehicle, to ensure that the system behaviour is predictable and no safety issues arise.

Towards this, several simulation frameworks have been developed to validate their corresponding components of the vehicle while approximating the other parts of the system.
For example, control systems can be developed using MATLAB~\cite{matlab} to evaluate their stability during operation.
However, the vehicle's state is often only estimated by using the corresponding kinematic and dynamic models instead having an accurate vehicle simulation.
Other examples are discrete event simulators, like the ns-3~\cite{ns3}, VANET Toolbox~\cite{vanet-toolbox} and OMNeT++~\cite{omnet}, which can be used to simulate and analyze vehicular networks.
Nevertheless, they require external frameworks such as the SUMO traffic simulator~\cite{sumo} to approximate each vehicle's movement, avoiding to simulate each vehicle's navigation and control systems by itself.
Finally, the Robot Operating System (ROS)~\cite{ros} provides implementations of algorithms for mapping, localization, navigation and more to create complex navigation systems.
It can be used together with robot simulation tools like Gazebo~\cite{gazebo}, Carla~\cite{carla}, or LG SVL~\cite{lgsvl}, which provide a graphics and physics engine to simulate the environment, the vehicles, and their sensors for various scenarios.
However, ROS was not originally designed for multi-robot scenarios.
For this purpose, ROS2 was released, which includes many architectural improvements that can be used to design more complex navigation systems and connect to multi-robot simulations.
Nevertheless, it does not provide the level of detail for simulating communication networks and has limited design tools for control systems in comparison to MATLAB.
In general, such practices result in researchers using an oversimplified approximation of system components from other domains and a lack of consistent methods for evaluating various scenarios with reasonable effort.


In order to tackle the aforementioned issues, in this work, we develop a new framework, which achieves such an integration over state-of-the-art tools, i.e., ROS2, OMNeT++ and MATLAB, for the simulation of autonomous driving scenarios.
The framework keeps the modular architecture of each simulation tool so that new components can be implemented in their corresponding development environment, while the evaluation features the complete system, including the robot, communication and control system simulation.

\noindent\textbf{Our Contributions in a Nutshell:}
\begin{itemize}
    \item We integrate ROS2, OMNeT++ and MATLAB to create an integrated simulation framework named \uline{Au}tonomous \uline{Na}vigation that is capable to simulate various cooperative autonomous driving scenarios (see Section~\ref{sec:framework-overview}).
    \item To demonstrate the applicability, we deploy a platooning scenario using cooperative adaptive cruise control in the framework, where the ETSI ITS-G5 architecture is simulated for the communication (see Section~\ref{sec:case-study-platooning}).
    \item Through extensive experiments, we showcase the effects of using the communication standard and compare the control performance between a theoretical evaluation and practical simulation (see Section~\ref{sec:evaluation}).
\end{itemize}

The framework is publicly available at Github:  (\url{https://github.com/tu-dortmund-ls12-rt/AuNa}).


\section{Background}

This section presents the targeted cooperative driving scenarios and an overview of the currently available simulation frameworks for cooperative multi-robot simulations.

\subsection{Cooperative Driving Scenarios}

The CAR 2 CAR Communication Consortium defined three phases for cooperative driving scenarios for future autonomous vehicles, which are referred to as Cooperative Intelligent Transport Systems (C-ITS)~\cite{c-its,c-its-car-2-car}.
The first phase includes warning signals for intersections, traffic jams or other road users which are transmitted between vehicles to enhance foresighted driving.
The second phase improves the service quality by sharing perception and awareness messages to implement overtaking warnings, cooperative adaptive cruise control or detailed warning signals.
The third phase includes cooperative driving scenarios, such as platooning, lane merging and overtaking.
Furthermore, the vehicles not only share information messages but also negotiate the outcome during navigation for the third phase.

A successful implementation of these scenarios increases fuel efficiency and road capacity, as well as the navigation to improve the driving experience.
Furthermore, the safety is increased, as each vehicle maintains an appropriate safety distance during operation and traffic accidents are avoided.

In this paper, we simulate a vehicle platoon by implementing cooperative adaptive cruise control (CACC)~\cite{cacc-controller}, which is part of the second phase.
For this, the vehicles communicate to exchange their state and try to maintain an appropriate safety distance to the leading vehicle.
As a result, each follower does not have to determine the state of other vehicles using its own sensors, improving its perception of the environment.
Instead, the communication and control systems need to be reliable to successfully implement CACC.

\subsection{Cooperative Communication}

For the communication of C-ITS, the ETSI ITS-G5 communication standard~\cite{etsi-its-g5-standard} is defined.
Artery, which is an extension of OMNeT++, features an implementation of the ETSI ITS-G5 communication architecture to simulate the communication between the onboard units (OBU) of the vehicles and roadside units (RSU) in the infrastructure.

Artery includes several blocks, consisting of the physical layer and MAC layer of Veins~\cite{veins}, the transport and networking layers of Vanetza, and a configurable application layer that is connected via a managing middleware.
It can be used to modularly implement applications, such as platooning, and connect other frameworks to OMNeT++ by integrating it into the corresponding layers of Artery.

For C-ITS scenarios, Artery provides services in the application layers, which implement the communication rules for the message transfer.
Natively, it implements the Cooperative Awareness (CA) basic service~\cite{cam-standard} of the ETSI ITS-G5 standard, which automatically broadcasts Cooperative Awareness Messages (CAM) to communicate with other road users.

\subsection{Related Work}

This section provides an overview of available tools from robotics, communication and control domains and their shortcomings for cooperative driving scenarios.

\noindent\textbf{Robotics:} 
The Robot Operating System (ROS)~\cite{ros} is a collection of tools and libraries for creating robot applications, such as navigation systems for autonomous vehicles.
ROS implements components for robot systems as nodes, such as algorithms for localization, mapping, and navigation.
Additionally, these nodes can communicate via topics, which are used as message busses for inter-node communication.
For example, a node can publish a message to a topic, which is then broadcast to all nodes that subscribe to that topic.
ROS provides a modular approach for implementing robot applications and many packages are available that provide nodes and tools to design complex robot systems.

However, ROS was originally designed for single robot systems with ideal communication, without considering real-time constraints and embedded systems.
Although some packages try to enable these features, they are built on the limited architecture of ROS.
This could lead to further limitations when developing future autonomous driving systems, which would have to be corrected by the application itself.

\noindent\textbf{Communication:} OMNeT++~\cite{omnet} is an extensible, modular simulation library and framework for building network simulators.
It is a discrete event simulator for different network architectures with wired or wireless communication.
However, OMNeT++ requires external tools to simulate vehicles and update the simulation.
For example, the Simulation of Urban Mobility (SUMO) tool simulates the traffic and provides OMNeT++ with the vehicle states.
Nevertheless, SUMO does not simulate the vehicle sensors or navigation systems, but only approximates their movement in traffic scenarios.
Other discrete event simulators like ns-3 or the VANET Toolbox suffer from the same problem, requiring external data or simulators to provide the vehicle data for the simulation.

\noindent\textbf{Control:} Control systems can be evaluated with MATLAB and Simulink.
MATLAB is a numeric computing environment designed to create and evaluate algorithms for areas such as signal processing, machine learning and robotics.
Simulink is a MATLAB package for control system design and evaluation.
It provides predefined blocks which can be arranged and connected to create complex control systems and can be evaluated using artificial or recorded input signals.
However, the vehicle data is often approximated by defining the corresponding kinematic and dynamic models for the vehicle movement and creating artificial signals with Gaussian noise for the sensor data.
Alternatively, the recording of signals is time consuming and expensive.

\noindent\textbf{Integrated Solutions:} The COPADRIVe framework~\cite{copadrive} integrates ROS and OMNeT++ to simulate the robot and their navigation systems as well as the communication.
However, it is specifically designed to evaluate platooning scenarios, which does not provide the flexibility to evaluate other cooperative driving scenarios for future applications.


\section{Framework Overview}
\label{sec:framework-overview}

For the simulation of cooperative driving scenarios, the following four main components need to be implemented and integrated to interact with each other:

\begin{itemize}
    \item \textbf{Robot Simulation}: The robot simulation is responsible for creating a virtual environment and the robots.
    It should include a graphics engine for evaluation and a physically accurate simulation to provide the navigation systems and other simulations with vehicle controls, sensor data, and the current state of the simulation.
    \item \textbf{Communication Simulation}: The simulation of the communication should implement the communication standard for cooperative scenarios to simulate effects such as delays and package loss.
    Additionally, it needs to be synchronized with the robot simulation and each robot needs to interact with its communication module.
    \item \textbf{Control Simulation}: The control simulation should implement the control systems for each vehicle.
    Furthermore, it should provide a modular architecture to design complex systems and provide the functionalities to connect them to the navigation systems of each robot.
    \item \textbf{Navigation Systems}: Each robot requires a navigation system that receives the data from all simulations and processes it to create a map of the environment, localize, and navigate itself in the scenario.
\end{itemize}

For multi-robot simulations, all robots should be included in the same robot simulation simultaneously, while their navigation systems are independent of each other.
The communication between the robots for cooperative scenarios should happen through the communication simulation.

Our framework is based on ROS2-Foxy~\cite{ros2} to implement the robot systems and integrate the tools into a framework.
ROS2 provides a lightweight and more robust version than ROS, improves upon already present aspects, and provides new functionalities for multi-robot setups.
For example, the communication between nodes and topics is implemented using Data Distribution Services (DDS), which provide more efficient and robust communication and are suitable for autonomous vehicles and embedded systems.

In general, ROS2 nodes can implement any algorithm to process the received data and transmit their results to other nodes.
They can be implemented in C++ and Python and can be integrated into other tools by importing the corresponding libraries.
For the cooperative driving simulation, each tool should implement nodes to interact with each other.

The following technical \textbf{challenges} must be overcome for the integration of all tools into the framework and to create a complete and modular multi-robot simulation:

\begin{itemize}
    \item \textbf{C1 (Flexibility \& Efficiency)}: The simulation tools should efficiently communicate to synchronize their entities across the simulations without manual adjustments.
    Additionally, the simulation should be \emph{configurable} by the user to adjust the scenario easily across all tools.
    For example, the simulation of additional robots should only require to specify the needed number of robots instead of adding them individually.
    \item \textbf{C2 (Event Management)}: Each simulation tool should \emph{directly manage} its simulation by itself and not be controlled by other simulation tools.
    For example, OMNeT++, as a discrete event simulator, should only receive the data about the current state of the robot simulation and update its simulation environment itself.
    \item \textbf{C3 (Event Synchronization)}: All simulators should \emph{efficiently synchronize} the generated events under a global view of a synchronized clock without affecting the other parts of the simulation.
    Specifically, the communication simulation should synchronize itself with the robot simulation so that the environment represents the most recently received simulation state.
    \item \textbf{C4 (Modular Preservation)}: The framework should \emph{preserve the modular architecture} of each tool and should not require the user to modify core parts of the architecture.
    By doing so, updates of different tools can still be integrated in the framework with a minimum effort, e.g., the integration of new applications and system designs should not require to adapt the underlying layers of the system architecture.
\end{itemize}

\subsection{COPADRIVe Integration and Limitation}
\label{sec:COPADrive-Challenges}

To the best of our knowledge, COPADRIVe~\cite{copadrive} is the first integrated simulator capable of simulating a multi-robot scenario for cooperative autonomous driving.
However, the integration is either limited by its tools or does not make efficient use of the underlying modular architectures.
Therefore, it does not fully overcome the above challenges.

For COPADRIVe, ROS launch files that start the required nodes for the navigation systems and robot simulation have to be manually adjusted to extend and configure the scenario, which limits the flexibility that is required for \uline{C1}.

COPADRIVe uses OMNeT++ and Artery to simulate the V2X communication.
However, COPADRIVe does not efficiently synchronize the robot simulation and communication simulation to address \uline{C1} ,\uline{C2} and \uline{C3}.
Furthermore, the architecture of Artery is adjusted to only support the platooning scenario, so that additional scenarios cannot be implemented easily, which is required for \uline{C4}.

For the control systems, COPADRIVe includes a PID-controller, but does not make use of MATLAB and Simulink.
In general, this approach can be used for simple control systems, but limits the flexibility for \uline{C1} to easily implement complex control system designs, such as model-predictive or machine-learning based methods for future scenarios.

In conclusion, while COPADRIVe successfully integrates these tools for the platooning scenario, it unfortunately does not fully overcome the previously mentioned challenges.
The following sections introduce our approaches to address these issues and overcome the previously mentioned challenges.

\subsection{Framework Integration}
\label{subsec:framework_integration}

We introduce the adopted methods for simulating multi-robot scenarios in a step-wise manner, including every simulation component that is introduced in Section \ref{sec:framework-overview}.

\paragraph{Robot Simulation} The robot simulation should provide an environment in which multiple robots can be dynamically spawned.
Each robot needs a unique name or index that will be used across all tools.
The simulation should provide the current simulation time and the state of the currently simulated robots.
For each robot, a connection needs to be established via ROS2 nodes, which publishes the generated sensor data and enables the control of the robots.

\begin{figure}[t]
	\centering
	\includegraphics[width=0.55\columnwidth]{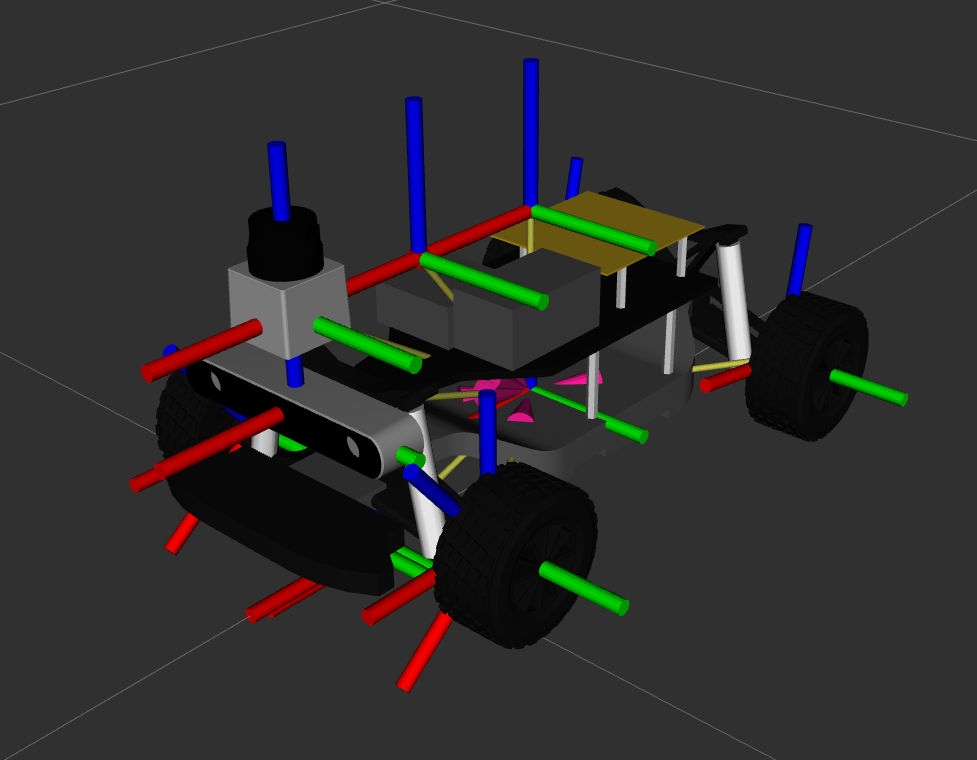}
	\caption{Robot model including the robot link reference frames (XYZ-axis) and their transformations (yellow)}
	\label{fig:rviz_tf}
\end{figure}

For the robot simulation, Gazebo provides plugins that implement the sensors and the robot controls.
In addition, it supports functionalities that publish the current state and time of the simulation.
Each robot is given a unique name or index by adding a namespace to the Gazebo plugins, so that each node and topic corresponds to its robot.
In addition, the robots publish their transformations, which are used to determine their current local and global positions.
In general, ROS2 does not include namespaces for these transformations.
However, to simulate multi-robot scenarios, the namespace is appended to each transformation link, so that they are unique to their corresponding robot.
This allows to differentiate and visualize all robots under a global transformation tree.
Furthermore, ROS2 includes extended launch files, which are written in Python, so that the number of spawned robots can be adjusted by the user during startup.
Therefore, \uline{C1} is addressed to efficiently scale the simulation.

\paragraph{Communication Simulation} OMNeT++ and Artery are used for the communication architecture in this framework.
The following steps cover how we address the challenges to create an efficient and modular framework.

The first step is the connection and synchronization between OMNeT++ and the other simulations using ROS2 nodes.
We implement a custom scheduler within OMNeT++ that runs the execution of OMNeT++ events, the synchronization between OMNeT++ and Gazebo, as well as the updates between the navigation system in ROS2 and communication module in OMNeT++ of each robot.
For each planned event, the scheduler first processes the functions of the ROS2 node to receive the current simulation time and update the robot states.
After that, it schedules an OMNeT++ event if the event time is earlier than the current simulation time of the robot simulation.
Otherwise, the scheduler keeps updating until enough time has passed.
This approach solves \uline{C1}, \uline{C2}, and \uline{C3} by implementing an efficient synchronization method based on the event-driven architecture of OMNeT++.


The second step is the creation of communication modules for each robot.
We use the implemented node to request and receive the current state of the Gazebo simulation.
For each robot, OMNeT++ checks whether the corresponding module is already present or not and spawns new modules as necessary.
As a result, a corresponding OMNeT++ module is created for each robot system on the fly without over provisioning, therefore \uline{C1} is solved in an efficient way.

The final step is the integration of the communication modules to communicate with their corresponding navigation systems.
We further extend the mobility module in Artery to receive the currently estimated state of the robot, which includes the position, velocity, acceleration, heading and yaw rate.
As defined by the architecture of Artery, the mobility module automatically notifies the other components, such as the middleware and vehicle data provider, so that each of them is updated accordingly and the communication can take place.
For additional applications such as the platooning scenario, we implement an additional service into the application layer, which forwards the messages of the leading vehicle to the navigation system.
This approach preserves the modular architecture of Artery so that future applications can be added independently of each other and the architecture is used to its full extent, addressing \uline{C1} and \uline{C4}.

\begin{figure}[t]
	\centering
	\includegraphics[width=0.4\textwidth]{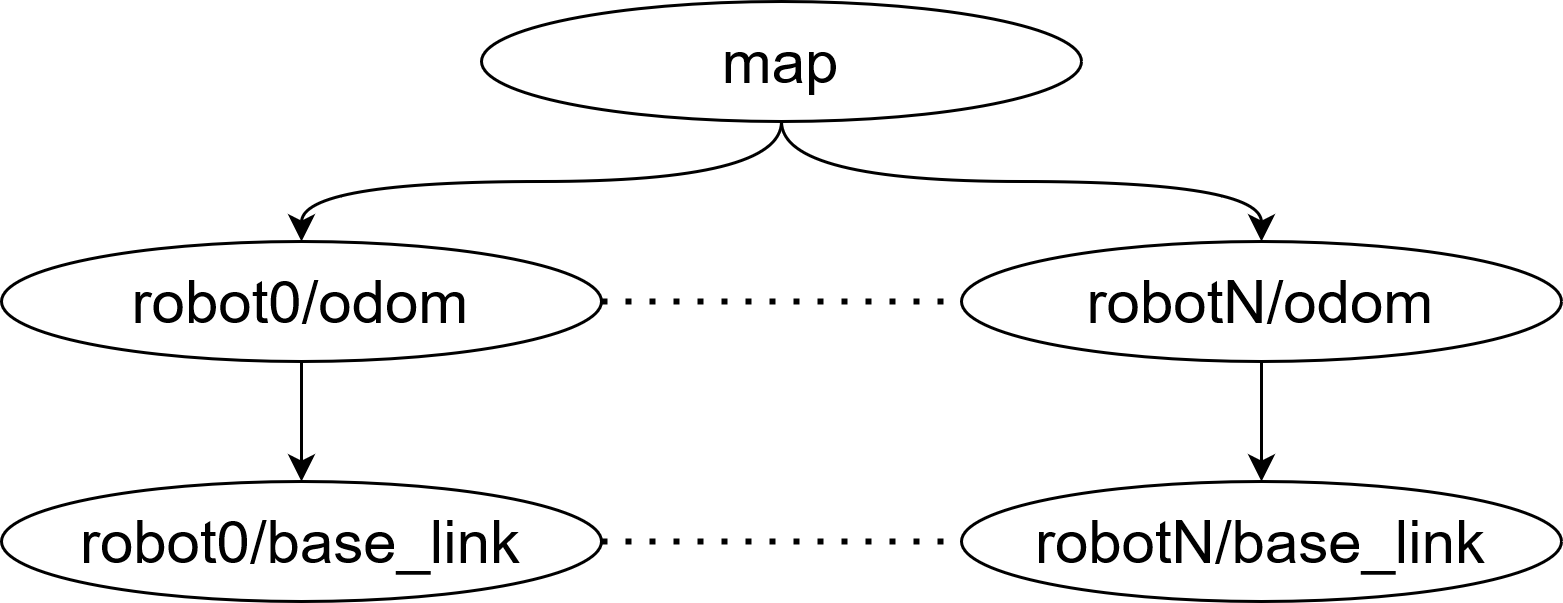}
	\caption{Multi-robot setup transformation tree}
	\label{fig:multi_transformation_tree}
\end{figure}

\paragraph{Control Simulation}

Instead of implementing the controllers directly in code, MATLAB and Simulink are integrated into the framework, as they provide a complete feature set for designing control systems of any complexity.
The 2022a release of MATLAB and Simulink provides ROS2 support with the ROS Toolbox.
Therefore, control systems can be connected by replacing the inputs and outputs with the corresponding topics of the navigation system.
Using the Parallel Computing Toolbox, a common control design can be launched for each individual robot by only specifying the currently simulated robots, so that \uline{C1} and \uline{C4} are addressed.

\paragraph{Navigation Systems}

The robots in the simulation require a navigation system to autonomously reach a target location.
To this end, we consider the Nav2 package~\cite{nav2} in ROS2, which provides a collection of algorithms for localization, mapping, and navigation that are designed for dynamic environments.
As in the robot simulation, we use extended launch files to simultaneously launch a corresponding navigation system for each spawned robot and differentiate them using namespaces.
In addition, we also include the namespaces for the transformations of the nodes, as mentioned for the robot simulation.
This solution overcomes \uline{C1} and efficiently utilizes the architecture of ROS2.

\subsection{Integration Summary}
\label{sec:integration-summary}

Integrating multiple tools into a unified framework poses the aforementioned challenges and requires to adapt the components in a way which keeps the core functionalities while having a seamless integration.
The presented solution solves these problems and does not require the user to further adapt the framework, but to build upon it and develop the applications directly, having an efficient and modular foundation to work with.
This enhances the development process to include the functionalities of each tool while integrating them for a complete cooperative multi-robot application.


\section{Case Study -- Platooning Scenario}
\label{sec:case-study-platooning}

To demonstrate the applicability, a platooning scenario under cooperative adaptive cruise control (CACC) is deployed on the framework.
In the following subsections, we present the details of the robot systems and the implementation of the platooning controller and communication service.

\begin{figure}[t]
    \centering
    \includegraphics[width=0.65\columnwidth]{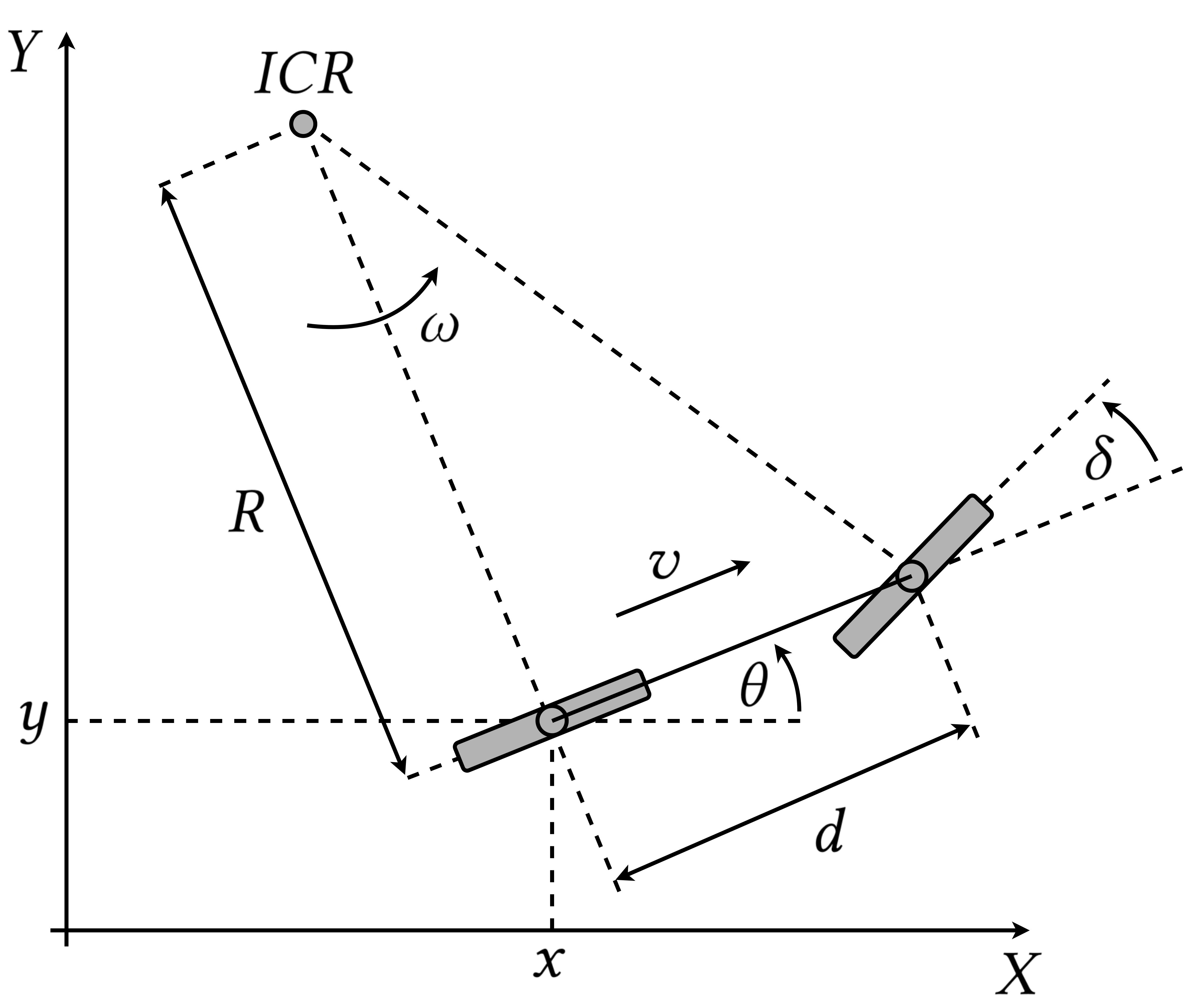}
    \caption{Kinematic bicycle (single-track) model of a vehicle.}
    \label{fig:bicycle_drive}
\end{figure}

\subsection{Robot Simulation}

The first part is the Gazebo simulation.
The robot model is shown in Fig. \ref{fig:rviz_tf} and the simulated environment in Fig. \ref{fig:racetrack_track}.
The robots are equipped with wheel sensors to track their movement and a LIDAR sensor to sense the environment.
They can be controlled via their Ackermann-drive, which requires an input velocity and steering angle during navigation.
The wheel sensors, LIDAR, and the drive are implemented via ROS2 Gazebo plugins and the transformations of the robot are modified as described in Section \ref{subsec:framework_integration}.

\subsection{Navigation System}

An autonomous navigation system is implemented for the robot, which includes multiple state variables.
The current robot pose $x_t=(x,y,\theta)^T$ is defined as the location of the robot, including its $x$ and $y$ coordinates, as well as its orientation $\theta$.
The robot movement is defined by its control vector $u_t=(v,\omega)^T$, which includes its velocity $v$ and yaw rate $\omega$.
The control vector can be calculated using the wheel sensors, as they returned the traversed distance for each time step.
The LIDAR sensor returns a measurement of its surroundings, which is an array of distances that can be used to determine the locations of objects in the environment.

\begin{figure}[t]
	\centering
    \begin{subfigure}{0.35\textwidth}
        \includegraphics[width=\textwidth]{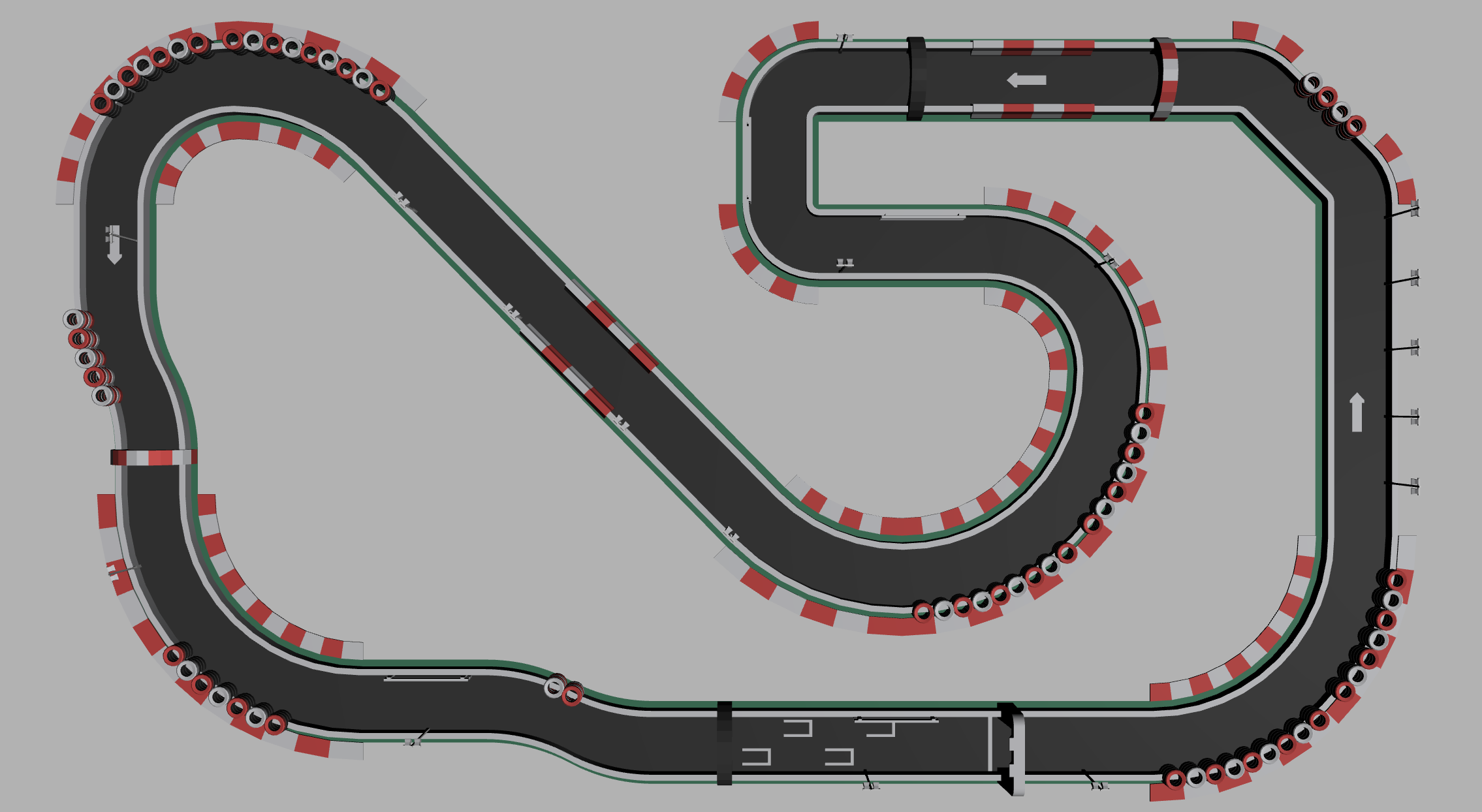}
        \caption{Simulated racetrack in Gazebo}
        \label{fig:racetrack_track}
    \end{subfigure}
    \begin{subfigure}{0.35\textwidth}
        \includegraphics[width=\textwidth]{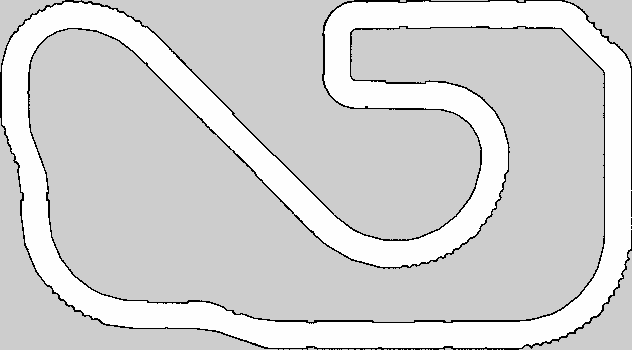}
        \caption{Racetrack map using the SLAM-Toolbox package}
        \label{fig:racetrack_map}
    \end{subfigure}
    \caption{Simulated environment and resulting map}
    \label{fig:racetrack}
\end{figure}

During operation, the robot processes its sensor data to track its pose $x_t$ and generate a map $m$ of the environment.
The navigation system uses a grid map that represents the environment using a grid of cells, which represent the occupied and free space in the environment.
The resulting map of the racetrack is shown in Fig. \ref{fig:racetrack_map}.

During navigation, each robot keeps track of its pose using the odometry data of its Ackermann-drive.
For the other components, the movement of the robot is approximated using the bicycle drive~\cite{kinematic-drive} that is shown in Fig. \ref{fig:bicycle_drive}.
Given its wheelbase $d$,  steering angle $\delta$ and the control vector, the new location of the robot is given by the following equations:
\begin{equation}
	x' = x + v \cdot \cos(\theta) \cdot\Delta t \;\;\;\mbox{ and }\;\;\;
	y' = y + v \cdot \sin(\theta) \cdot\Delta t
\end{equation}

The radius of the robot movement and the steering angle are given by the following equations:
\begin{equation}
	\frac{v}{\omega} = R
\;\;\;\mbox{ and }\;\;\;
	\tan(\delta) = \frac{d}{R}
\end{equation}

These are solved for $\omega$ to calculate the new robot heading:
\begin{equation}
	\theta' = \theta + \frac{v \cdot \tan(\delta)}{d} \cdot \Delta t
\end{equation}

\begin{figure}[t]
    \centering
    \includegraphics[width=0.475\textwidth]{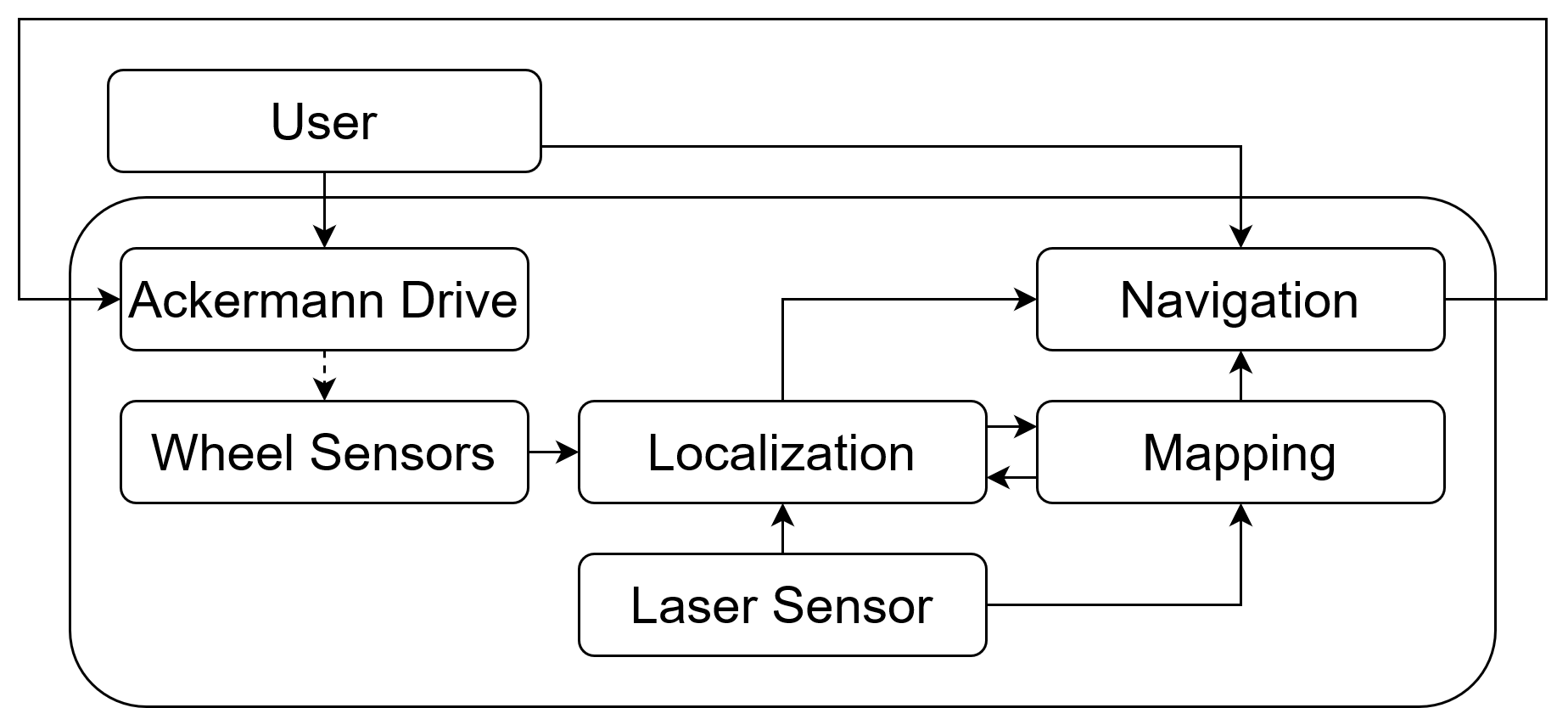}
    \caption{Robot navigation system overview}
    \label{fig:robot_navigation_system_overview}
\end{figure}

For the evaluation, several packages are used to implement the components of the navigation system in Fig. \ref{fig:robot_navigation_system_overview}.

\begin{itemize}
    \item The SLAM-Toolbox package~\cite{SLAMToolbox} maps the environment, as shown in Fig. \ref{fig:racetrack_map} for the racetrack.
    \item The AMCL package of Nav2 provides an implementation of Augmented Monte Carlo Localization~\cite{amcl-paper} for global localization on the previously generated map.
    \item The Nav2 package includes costmaps~\cite{Costmap2} to take dynamic obstacles in the environment into account.
    \item The Nav2 package implements A*~\cite{astar-book} for global navigation and DWA~\cite{dwa-paper} for local navigation.
\end{itemize}

The resulting robot is capable of navigating itself in dynamic environments and provides the other simulations with the required data to communicate with other robots and implement the control systems for the scenarios.

\subsection{Communication Simulation}

For each robot, a node is implemented such that it collects and forwards the current state of the robot to Artery and OMNeT++, including the position, velocity, acceleration, heading and yaw rate.
The communication for platooning is implemented by adding a new platooning service to Artery.
Through the already implemented CA basic service, the vehicles exchange CAMs that contain the information about the other's state.
The platooning service of Vehicle $i$ only considers the CAMs from its leading Vehicle $i-1$, for $i \geq 2$ and forwards them to the navigation system of the robot.
Afterwards, the navigation system receives and processes the data directly or through its control systems.

\subsection{Control Simulation}

For platooning, a controller is required to calculate the appropriate acceleration and yaw rate to maintain a stable safety distance during operation.
In this paper, the controller designed in~\cite{cacc-controller} is implemented in MATLAB and Simulink and connected using the ROS and Parallel Computing Toolbox.
For completeness, we sketch their controller here.

The controller is responsible for the longitudinal and lateral control implements a time-gap spacing policy with a standstill distance of $r_i~ meters$ and a time gap of $h_i~seconds$.
It is designed for non-zero positive velocities and takes the velocity and longitudinal and lateral acceleration into account.
Furthermore, it avoids cutting corners by implementing an extended look-ahead approach and features a time-gap spacing policy instead of a constant spacing.

For longitudinal control, it defines the error $e_i$ as the difference between the real distance $d_i$ and the desired distance $d_{r,i}$ between the positions $p_i$ and $p_{i-1}$:
\begin{equation}
	e_i = d_i-d_{r,i} = (p_{i-1} - p_i) - d_{r,i}
\end{equation}
\begin{equation}
	d_{r,i}= 
	\begin{bmatrix}
		d_{rx,i}\\
		d_{ry,i}
	\end{bmatrix}
	= (r_i+h_i\cdot v_i)
	\begin{bmatrix}
		\cos\theta_i\\
		\sin\theta_i
	\end{bmatrix}
\end{equation}

However, using this error would lead to the vehicle cutting corners, as it always directly targets the desired position, without taking the leading vehicle's heading into account.

The extended look-ahead approach in~\cite{cacc-controller} fixes this issue by defining a new look-ahead point which includes the leading vehicle's heading in the extension vector $s_{i-1}$.
It applies the following error term for the case of $\omega_{i-1}\not=0$:
\begin{equation}
	e_i = (p_{i-1}+s_{i-1})-(p_i+r_i+h_i \cdot v_i)
\end{equation}

The controller outputs consists of the acceleration and yaw rate of a unicycle drive robot.
It needs to be converted to the corresponding output velocity and steering angle of the Ackermann-drive.
The kinematic model of a unicycle drive is described by the following equations:
\begin{equation}
	\dot{x}_i = v_i\cdot \cos(\theta_i),\;\;\;\;
	\dot{y}_i = v_i\cdot \sin(\theta_i), 
\end{equation}
\begin{equation}
	\dot{v}_i = a_i, \;\;\;\;
	\dot{\theta}_i = \omega_i
\end{equation}

The output velocity corresponds to the integral of the acceleration, which can be implemented using an Integrator in Simulink.
Additionally, it is filtered to only return positive values and turned off in case the leading vehicle is standing still or moves slower than a threshold value.
Next, the following equation is used to convert the yaw rate of the robot to the steering angle of the bicycle-drive model:
\begin{equation}
	\delta = atan\left(\frac{\omega \cdot d}{v}\right)
\end{equation}

The controller in~\cite{cacc-controller} is configured by four parameters, including the gains for longitudinal and lateral control as well as the standstill distance and time-gap.


\section{Evaluation}
\label{sec:evaluation}

Using the previously introduced robot system and the framework, the following aspects were evaluated:

\begin{itemize}
	\item Compare an ideal channel that is implemented by a ROS2 topic to the communication simulation via Artery and the ETSI ITS-G5 architecture.
	Highlight the effects of the communication on the received data.
	\item Compare the theoretical controller performance using synthetic input signals and approximated vehicles to the controller performance when used in the framework, which accurately simulates the robot movement and includes the communication effects.
	This highlights the difference between the performance using approximations under ideal conditions and an accurate simulation.
\end{itemize}

Head-to-head comparisons between COPADRIVe and our modularly integrated simulation framework would be able to demonstrate how much improvement we have achieved.
However, it would require to port their robot simulation environment, navigation, and control systems to ROS2 and would include the pitfalls of their integration pointed out in Section~\ref{sec:COPADrive-Challenges}.
Therefore, the evaluation only showcases scenarios that are simulated using our framework.

\subsection{Evaluation Setup}

The evaluation was conducted on a system using Ubuntu 20.04.
It includes an AMD 5900x processor with 12 cores that run at a base frequency of $3.7GHz$ and a boost clock of $4.8GHz$.
Simultaneous hyperthreading is enabled, resulting in 24 logical cores.
The system has $32GB$ of DDR4 memory that runs at $3600MHz$ and an AMD Radeon RX 6800XT graphics card with $16GB$ of memory.

The platoon consists of four robots, which are simulated in Gazebo and feature the sensors and navigation systems, introduced in Section \ref{sec:case-study-platooning}.
The communication is implemented using Artery and OMNeT++, while the control systems in Section \ref{sec:case-study-platooning} are integrated using MATLAB and Simulink.

\subsection{Communication Evaluation}

\begin{figure}[t]
    \centering
    \includegraphics[width=0.4\textwidth]{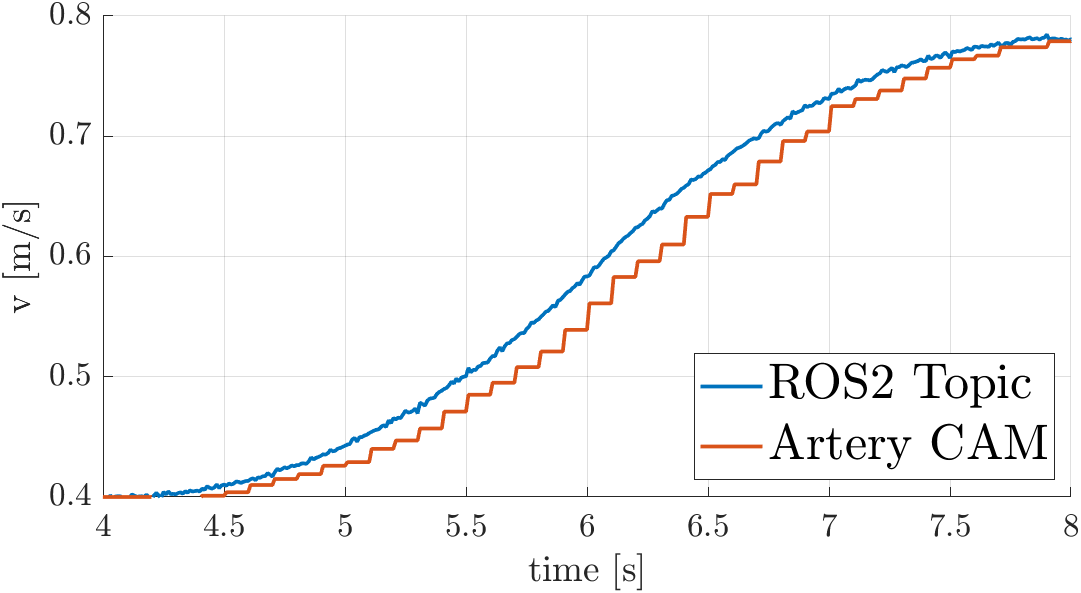}
    \caption{Received velocity data by ideal ROS2 connection and ETSI ITS-G5 architecture during platooning}
    \label{fig:evaluation_communication}
\end{figure}

Fig. \ref{fig:evaluation_communication} shows the communication between the leading vehicle and the follower when transmitting the leading vehicles's velocity.
In comparison to an ideal connection via a ROS2 topic, the transmission of CAMs introduces a delay, as well as a lower frequency and lower data resolution.
The delay is quite consistent and ranges from $0.1-0.2$~seconds.
There are two main factors which influence the delay, the processing time to transmit and receive CAMs as well as the transmission delay between the vehicles.
The frequency is controlled by the rules that are defined by the ETSI ITS-G5 standard, which limits the transmission interval to be between $0.1-1$~seconds.
The lowered data resolution is due to the message format of CAMs, as it uses a 16-bit integer instead of a 64-bit float that is used by the ROS2 topic.

The framework successfully integrates ROS2 and Artery to simultaneously simulate the robots and the communication between the vehicles.
The effects, such as delays and a lower data resolution, are present and can have an influence on the other parts of the robot system.
Therefore, the capabilities of ROS2 and OMNeT++ can be applied to implement the systems and architectures and evaluate their interactions.

\subsection{Controller Evaluation}

\begin{figure}[t]
\centering
    \begin{subfigure}{0.40\textwidth}
        \includegraphics[width=.95\textwidth]{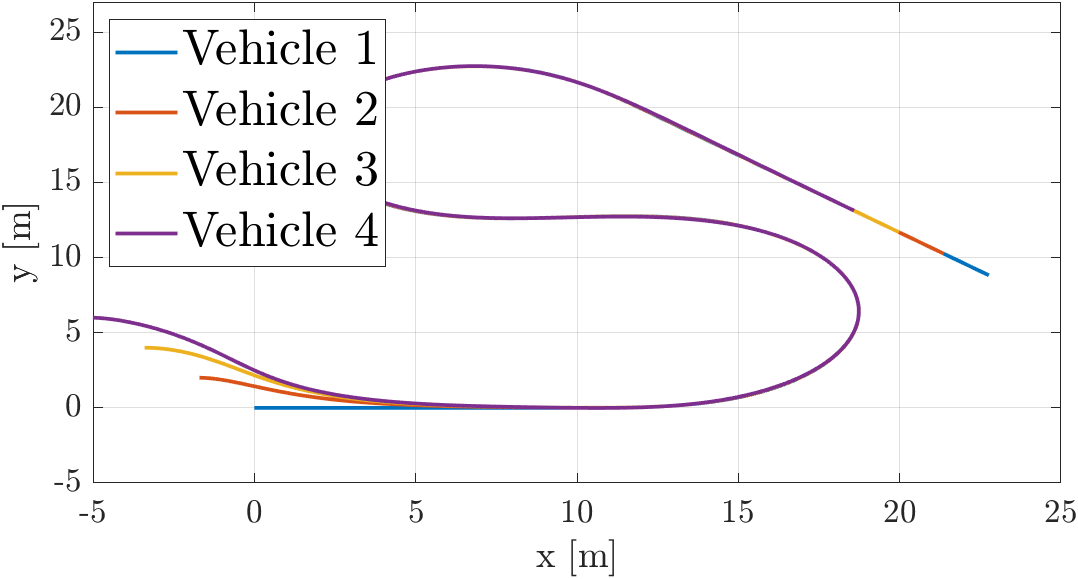}
        \caption{Theoretical scenario}
    \label{fig:evaluation_sim_scenario_position_a}
    \end{subfigure}
    \begin{subfigure}{0.40\textwidth}
        \includegraphics[width=.95\textwidth]{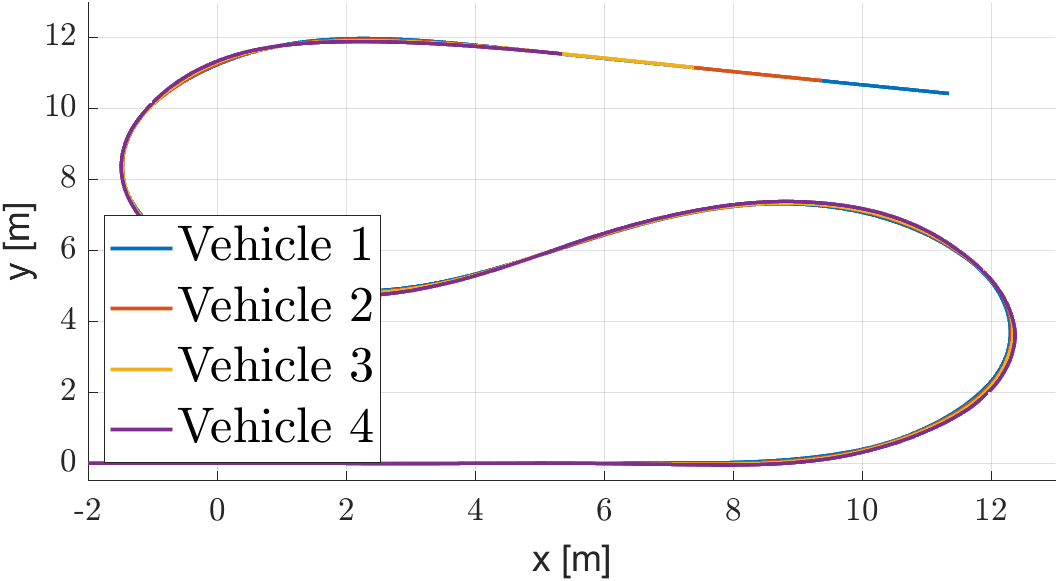}
        \caption{Simulated scenario}
    \label{fig:evaluation_sim_scenario_position_b}
    \end{subfigure}
    \caption{Trajectories between theory and simulation}
    \label{fig:evaluation_sim_scenario_position}
    \vspace{-0.5cm}
\end{figure}

The theoretical performance is evaluated completely in MATLAB, approximating the vehicles by using the unicycle-drive model and controlling the platoon leader by creating synthetic input signals via Simulink.
The outputs of each leading vehicle are directly forwarded to its corresponding follower so that the communication effects are not included.

\begin{figure}[t]
\centering
    \begin{subfigure}{0.40\textwidth}
        \includegraphics[width=0.95\textwidth]{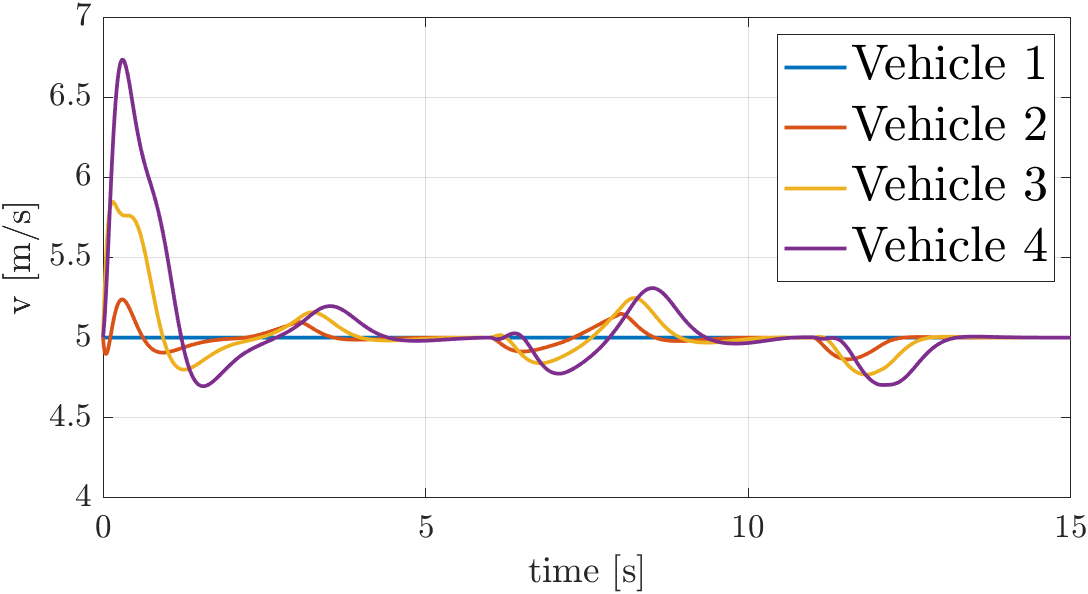}
        \caption{Theoretical scenario}
    \label{fig:evaluation_sim_scenario_velocity_a}
    \end{subfigure}
    \begin{subfigure}{0.40\textwidth}
        \includegraphics[width=0.95\textwidth]{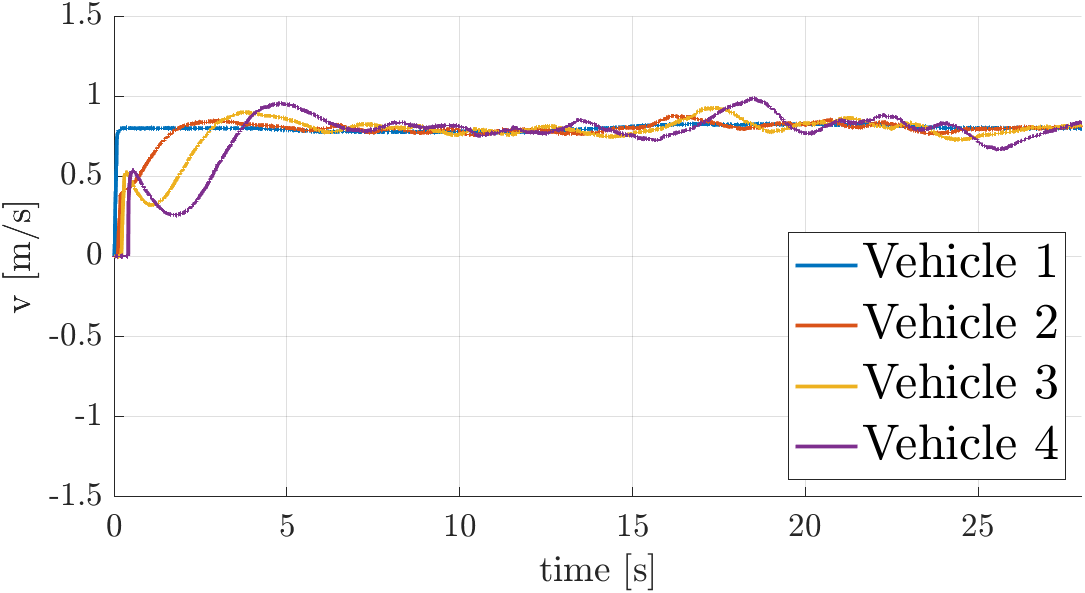}
        \caption{Simulated scenario}
    \label{fig:evaluation_sim_scenario_velocity_b}
    \end{subfigure}
    \caption{Velocity performance between theory and simulation}
    \label{fig:evaluation_sim_scenario_velocity}
    \vspace{-0.5cm}
\end{figure}

For the simulated scenario, the absolute position of the robots and their velocities are shown in Fig. \ref{fig:evaluation_sim_scenario_position} and \ref{fig:evaluation_sim_scenario_velocity}, where Vehicle $i+1$ follows Vehicle $i$ for $i\in\{1,2,3\}$.
The controller is configured to use the gains $(3.5, 3.5)$, standstill distance $1m$ and time gap $0.2s$ for the theoretical evaluation.

The theoretical performance of the controller produces almost perfect results, as the trajectories of the vehicles are nearly identical.
However, the controller is generally not string-stable, which can be observed in Fig. \ref{fig:evaluation_sim_scenario_velocity_a}.

For the framework evaluation, we use an empty world in Gazebo to drive similar curves to those of the theoretical evaluation.
The leader of the platoon, i.e., vehicle 1, is controlled by synthetic input signals.
As the map of the world is empty, the robot has to purely rely on its odometry data to determine its pose.
Therefore, the applied velocity is lowered to minimize the wheel slippage.
In addition, the controller is tuned for the robot model.
As a result, the gains are lowered to $(1.0, 1.0)$ and the time gap is increased to $1.0$~second.

As shown in Fig. \ref{fig:evaluation_sim_scenario_position_b} and \ref{fig:evaluation_sim_scenario_velocity_b}, the platoon performance in simulation is worse compared to the theoretical evaluation.
The trajectories do not overlap as much and the velocity of the vehicles includes additional errors.
This is due to the effects that each part has on the system, such as localization errors and communication effects.

In conclusion, the platooning scenario is successfully implemented, including the robot simulation in Gazebo, the navigation systems in ROS2, the communication in OMNeT++ and Artery, and the controller integration in MATLAB and Simulink.
The evaluation showcases that the theoretical and practical simulation produce different results.
Hence, the framework can be used to analyze the effects that each component has on the complete system.


\section{Conclusion}
This paper presents a simulation framework that integrates the state-of-the-art tools of robotics, communication and control systems, namely ROS2, Gazebo, OMNeT++, Artery, MATLAB, and Simulink.
It enables the simulation of cooperative autonomous driving scenarios and their required technologies.
As shown in the evaluation, a platooning scenario is successfully implemented by integrating the robot simulation, the navigation systems, the ETSI ITS-G5 communication standard, and a CACC controller.
Furthermore, the framework enables the evaluation of the effects that each component has on the scenario and other system components.

As the framework keeps the modular architecture of each simulation tool, new components from different domains can be integrated to extend the provided functionalities.
This includes other robot simulators such as Carla and LG SVL, more complex navigation systems like Autoware.Auto, 5G and 6G communication standards, as well as machine-learning based and model-predictive control systems.
Additionally, the framework can be used to design and implement the required components in their corresponding development environments and provides a consistent method for evaluating their performance in a complete system and scenario.


Overall, this work aims to accelerate the development of future cooperative autonomous driving technologies and support the cooperation between different research domains.
We plan an ongoing effort to keep the framework up-to-date, and evaluate more scenarios and the required components for cooperative autonomous driving in the future.

\addtolength{\textheight}{-12cm}   





\section*{Acknowledgment}
This is part of the "6G-Forschungs-Hubs; Plattform für zukünftige Kommunikationstechnologien und 6G" under the funding code 16KISK038, by the German Federal Ministry of Education and Research (BMBF).

{
\bibliographystyle{IEEEtran.bst}
\bibliography{main}
}
\end{document}